%% file: ARGUE.tex
\documentclass{article}
\usepackage{arxiv}  

\def \ARGUE {ARGUE}
\def \AAA {A\textsuperscript{3}}

\usepackage{graphicx}
\usepackage{subfig}
\usepackage{tikz}
\usepackage{pgfplots}
\usepackage{pgfplotstable}
\pgfplotsset{compat=1.16}

\usepackage{amsmath,amssymb,amsfonts}
\usepackage{cleveref}
\usepackage{enumitem}

\usepackage{makecell}
\usepackage{booktabs}
\usepackage{rotating}
\usepackage{tablefootnote}
\usepackage{multirow}
\usepackage{arydshln}

\usepackage{pgfkeys,pgfmath}  
\usepackage{siunitx}
\newcommand{\percincrease}[3][0]{%
    \pgfmathdivide{#3}{#2}%
    \pgfmathparse{\pgfmathresult-1}%
    \pgfmathmultiply{\pgfmathresult}{100}%
    \SI[round-mode=places,round-precision=#1]{\pgfmathresult}{\percent}
}%

\usepackage{abbr}
\usepackage{style}

\newcommand\copyrighttext{%
  \footnotesize \textcopyright 2022 IEEE. Personal use of this material is permitted. Permission from IEEE must be obtained for all other uses, in any current or future media, including reprinting/republishing this material for advertising or promotional purposes, creating new collective works, for resale or redistribution to servers or lists, or reuse of any copyrighted component of this work in other works.}
\newcommand\copyrightnotice{%
\begin{tikzpicture}[remember picture,overlay]
\node[anchor=south,yshift=2cm] at (current page.south) {\fbox{\parbox{\dimexpr\textwidth-\fboxsep-\fboxrule\relax}{\copyrighttext}}};
\end{tikzpicture}%
}

\title{Anomaly Detection by Recombining Gated Unsupervised Experts}

\author{
	Jan-Philipp Schulze \\
	Fraunhofer AISEC \\
	\texttt{jan-philipp.schulze\thanks{\texttt{@aisec.fraunhofer.de}}} \\
	\And
	Philip Sperl \\
	Fraunhofer AISEC \\
	\texttt{philip.sperl\footnotemark[\value{footnote}]} \\
	\And
	Konstantin B\"ottinger \\
	Fraunhofer AISEC \\
	\texttt{konstantin.boettinger\footnotemark[\value{footnote}]} \\
}

\begin{document}
\maketitle
\copyrightnotice

\begin{abstract}

Anomaly detection has been considered under several extents of prior knowledge.
Unsupervised methods do not require any labelled data, whereas semi-supervised methods leverage some known anomalies.
Inspired by mixture-of-experts models and the analysis of the hidden activations of neural networks, we introduce a novel data-driven anomaly detection method called ARGUE.
Our method is not only applicable to unsupervised and semi-supervised environments, but also profits from prior knowledge of self-supervised settings.
We designed ARGUE as a combination of dedicated expert networks, which specialise on parts of the input data.
For its final decision, ARGUE fuses the distributed knowledge across the expert systems using a gated mixture-of-experts architecture.
Our evaluation motivates that prior knowledge about the normal data distribution may be as valuable as known anomalies.


\end{abstract}

\section{Introduction}
In anomaly detection (AD), we look for inputs that differ from our training data.
Based on the setting, these anomalies may lead to e.g. security incidents, manufacturing errors or fraudulent behaviour.
In recent years, the superior performance of machine learning applications using deep learning (DL) has motivated active research in this area.
Here, relevant patterns in the input are detected by multi-layered neural networks (NNs).
AD poses a challenge to DL frameworks as usually only a clear notion of the normal behaviour exists.
Anomalies, however, do not follow a general pattern, but are merely defined by being different to the training data to some unknown extent.
In practice, there may be some prior knowledge that can be leveraged to boost the detection performance.
Popular types of prior knowledge are labelled anomalies as seen in semi-supervised AD (e.g. \cite{ruff_deep_2020,pang_deep_2019,sperl_activation_2021}), or known clusters of normal data as seen in out-of-distribution (OOD) problems (e.g. \cite{vyas_out--distribution_2018,hendrycks_baseline_2017,liang_enhancing_2018}).
We designed our AD method to work under diverse conditions found in anomaly detection problems.

In research, AD is usually seen as a monolithic problem where only a single notion of normal behaviour exists.
We broaden this view by introducing our novel AD method called \ARGUE{}.
In practice, the normal state may greatly shift: behavioural patterns differ between weekdays and weekends, factory plants consist of several machinery, and so forth.
In our research, we propose to split the notion of normal across several expert NNs that specialise on certain parts of the normal classes.
These splits are either determined by a clustering algorithm (unsupervised), by attributes within the input sample (self-supervised) or by manually labelled clusters (supervised).
Mixture-of-experts (ME) models \cite{jordan_hierarchical_1994} were introduced as a supervised method fusing the information of several supervised single-layered NNs, thus improving the overall classification performance.
We leverage this idea to improve unsupervised and semi-supervised AD.
Our novel architecture combines multiple expert deep NNs for more reliable AD performance even under noisy data sets.
Each of these expert networks is adapted to parts of the training data.
Based on this principle, we call our novel AD method \ARGUE{}: \underline{a}nomaly detection by \underline{r}ecombining \underline{g}ated \underline{u}nsupervised \underline{e}xperts.

\section{Related Work}
AD profits from a wide range of research across multiple domains.
There are methods applied to certain environments, e.g. federated systems \cite{nguyen_diot_2019}, certain data types, e.g. graphs \cite{bhatia_midas_2020}, under certain constraints, e.g. weakly-supervised \cite{pang_deep_2019} or semi-supervised \cite{ruff_deep_2020} environments.
One-class support vector machines (SVMs) \cite{scholkopf_support_2000} and Isolation Forest \cite{liu_isolation_2008} are among the most commonly known unsupervised AD methods. 

\paragraph{DL-based AD}
In recent years, progress has been made in DL-based AD \cite{pang_deep_2021,ruff_unifying_2021}.
DL methods can analyse high-dimensional inputs, but usually require large training data sets.
A widely applied idea are reconstruction-based approaches, e.g. using autoencoders (AEs) \cite{borghesi_anomaly_2019} or GANs \cite{schlegl_f-anogan_2019,akcay_ganomaly_2019,li_mad-gan_2019}
Other popular DL-based methods are e.g. Deep-SVDD \cite{ruff_deep_2018}, leveraging one-class classifiers, or DAGMM \cite{zong_deep_2018}, combining AEs and Gaussian mixture models (GMMs).
Coming from OOD detection, SSD \cite{sehwag_ssd_2020} explores self-supervised scenarios there, whereas we explore self-supervised scenarios in AD.
Recently, Sperl~et~al. \cite{sperl_activation_2021} and Raghuram~et~al.\cite{raghuram_general_2021} motivated that the activations of NNs can be used for AD.
In ARGUE, we improve this idea by analysing multiple NNs at once for shifts in their activation patterns.
This not only results in better detection performance, but also incorporates prior knowledge, which would have been unused otherwise.

\paragraph{Multi-Expert AD}
In contrast to all aforementioned methods, ARGUE fuses the information of multiple expert systems that are conditioned on parts of the normal training data.
ARGUE therefore profits from the various notions of normal, typically present in complex environments.
Combining the outputs of multiple SVMs on sub-classes of the data is an idea already explored in research \cite{wu_local_2007,theissler_multi-class_2017}.
In the scope of GAN-based AD, multiple concurrent generator-discriminator pair have a positive impact on the AD performance \cite{han_gan_2021}.
For OOD detection, Vyas~et~al. \cite{vyas_out--distribution_2018} proposed a method distributing the known classes among multiple NNs.
Instead, ARGUE is not only applicable to OOD, but also general semi- and unsupervised AD problems. 

\paragraph{Mixture-of-Expert Models}
ME models \cite{jordan_hierarchical_1994} combine multiple single-layered NN-based expert models to one overall decision system.
They were designed for supervised environments unlike unsupervised ones as found in AD.
Since their introduction, there has been active research on ME models \cite{yuksel_twenty_2012}.
The idea was transferred to k-nearest neighbour models \cite{milidiu_time-series_1999} and SVMs \cite{cao_support_2003} in the context of time-series forecast, or NN encoders for unsupervised domain adaptation \cite{guo_multi-source_2018}.
ME models were applied in the context of DL with thousands of expert systems \cite{shazeer_outrageously_2017}.
To the best of our knowledge, only two other authors have applied ME models in AD: Xia~et~al. \cite{xia_ensemble_2019} on system logs and Yu~et~al. on image data \cite{yu_mixture_2021}.
Although promising ideas, we see certain restrictions in the past research that we solve in ARGUE:
1) Data distribution: unlike the other methods, we divide the training data among the expert paths, even without ideal clusters known a~priori, 2) End-to-end neural architecture: past research introduced reconstruction errors as decision variable, yet we use an entirely DL-based architecture with significant increases in performance and flexibility, 3) Prior knowledge: to the best of our knowledge, we are the first to expand ME-based AD to self- and semi-supervised use-cases.

In summary, we make the following contributions:
\begin{enumerate}[topsep=0pt,noitemsep]
    \item We introduce ARGUE, a data-driven AD method applicable to unsupervised and semi-supervised settings fusing the knowledge of multiple expert NNs.
    \item We propose three strategies to distribute data among these expert NNs, and apply them to ten data sets.
    \item We evaluate ARGUE against ten AD methods and plan to open-source our code to support future research.
\end{enumerate}

\begin{table}[tb]
    \centering
        \caption{
        Types of prior knowledge considered in \ARGUE{}.
    }
    \label{tab:prior}
    \begin{tabular}{p{3.9cm} : p{3.9cm}}
         \hfil \textbf{A) Anomaly Labels} & \hfil \textbf{B) Normal Clusters} \\
         \midrule
        \begin{enumerate}[leftmargin=*,topsep=0pt,noitemsep]
            \item \emph{Unsuperv.:}
            no labels available.
            \item \emph{Semi-superv.:}
            labels of a few known anomalies avilable.
        \end{enumerate}
        &
        \begin{enumerate}[leftmargin=*,topsep=0pt,noitemsep]
            \item \emph{Unsuperv.:}
            no clusters known.
            \item \emph{Self-superv.:}
            clusters estimated from input.
            \item \emph{Superv.:}
            ideal clusters known.
        \end{enumerate}
    \end{tabular}
\end{table}

\section{Prerequisites}

As commonly used in literature, we define an anomaly as ``an observation that deviates considerably from some concept of normality'' \cite{ruff_unifying_2021} -- the characteristics of the deviation may not be known a~priori.
In AD, we discover samples that lie outside the regions of normality.
In \ARGUE{}, we consider two \emph{optional} types of prior knowledge: the anomaly label, i.e. if a data point is normal or anomalous, and the type of normality, i.e. if a data point belongs to the same class as another one.
Both information dimensions are often available in real-world scenarios, e.g. we might know that the current network packet belongs to a benign TCP connection.
\ARGUE{} is applicable in a wide range of scenarios as shown in \Cref{tab:prior}.

AD is different to OOD detection.
In \emph{OOD detection}, we discover inputs that deviate from the training samples of a given ML model, e.g. a classifier.
This implies supervised class labels, which are generally not available in AD.
\ARGUE{} is applicable OOD settings (B3 in \Cref{tab:prior}), but also works without known clusters (B1\&B2).
For an in-depth review about AD and its related field, we refer to the survey of Salehi~et~al. \cite{salehi_unified_2021}.
With \ARGUE{}, we provide a flexible \emph{AD method}, which easily integrates prior knowledge to boost the detection performance.

\subsection{Nomenclature}
We describe NNs as a function $f_\text{NN}(\vect{x}; \vectg{\theta}) = \hat{\vect{y}}$ approximating how the input $\vect{x}$ relates to the estimated output $\hat{\vect{y}}$ using the mapping parameters $\vectg{\theta}$.
In the following, we will use the abbreviation $f_\text{NN}: \vect{x} \mapsto \hat{\vect{y}}$.
Deep neural networks (DNNs) comprise multiple layers $f_{i, \text{DNN}}$, which are concatenated to the overall network $f_\text{DNN}=f_{L, \text{DNN}} \circ \ldots \circ f_{1, \text{DNN}}$.
When referring to NN, we usually mean DNN.
Each middle layer gives rise to the activations $\vect{h}_i$.
We denote the concatenation of multiple activations as $[\vect{h}_i]_i = [\vect{h}_0, \vect{h}_1, \ldots]$.

\subsection{Activation Analysis}
ARGUE transfers parts of the ideas of \AAA{} \cite{sperl_activation_2021} to our novel multi-expert AD method.
\AAA{} is a semi-supervised approach that comprises three NNs: the target, alarm, and anomaly network.
According to the authors' core assumption the activations $\vect{h}_i$ of the target network are different for samples, which it was trained on, and others, i.e. normal and anomalous ones.
The alarm network analyses these activation values, $f_\text{alarm}: [\vect{h}_i]_i \mapsto \hat{y}$.
While training, the anomaly network generates counterexamples from a Gaussian prior, $f_\text{anomaly}: \vect{x} \mapsto \tilde{\vect{x}} \sim \set{N}(0.5, 1)$.
The authors used AEs as target network.
AEs are a special type of NN where the input is reconstructed under the constraint of a small hidden dimension, $f_\text{AE}: \vect{x} \mapsto \hat{\vect{x}}$.
Whereas \AAA{} was only evaluated in semi-supervised settings, ARGUE also works in unsupervised ones and allows to use additional prior knowledge.
We analyse the activation patterns of several NNs at once and combine the detection results to the overall anomaly score.

\subsection{Mixture-of-Expert Models}
In ME models \cite{jordan_hierarchical_1994}, the decisions of multiple supervised expert NNs are combined to one overall output.
Each of the expert paths is trained on \emph{disjoint sets} of the data.
To combine the separate decisions, a gating mechanism is introduced, mapping its input to a probability distribution $\vect{p}=[p_j]_j$, e.g. a softmax-activated NN.
With multiple expert NNs and their scalar output $y_j$ the overall decision becomes:
\[
y_\text{out} = \sum_j p_j y_j = \trans{\vect{p}} \vect{y}.
\]
For ARGUE, we expand this idea to work in the unsupervised setting of AD.
Here, our gating network is a DNN analysing the activations of another network.

\section{ARGUE}
\begin{figure}[tb]
  \begin{center}
    \input{figs/high_level.tikz}
    \caption{
        ARGUE introduces an expert NN for each normal class, here learning to reconstruct the digits 0 and 1.
        The input ``0'' is known to the first expert NN thus the input is likely normal.
        All experts fail to reconstruct the anomalous sample, here: the digit ``2''.
        We show that the idea generalises to other applications of AD, where ideally separable classes might not exist.
    }
    \label{fig:highlevel}
  \end{center}
\end{figure}
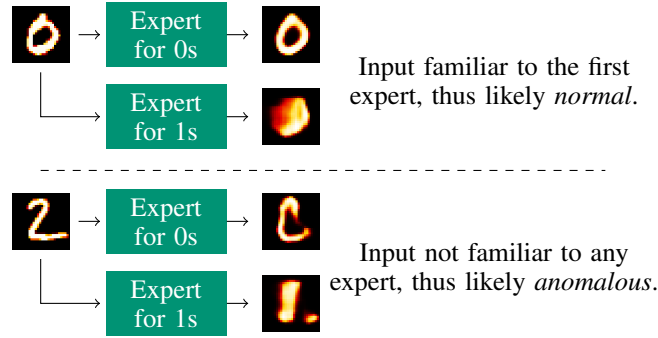

ARGUE builds on our core assumption:
\begin{quote}
    \label{assumption}
    Evaluating the activations $\vect{h}_{i, j}$ on layer $i$ of an expert neural network $f_j (\cdot)$, we observe special patterns that allow to distinguish between classes the network has been trained on, and unknown classes $y \notin \set{Y}_\text{train, j}$.
    Combining the knowledge of all expert neural networks, we can globally judge if a sample $\vect{x}$ belongs to a known class $y \in \set{Y}_\text{train} = \bigcup_j \set{Y}_\text{train, j}$.
\end{quote}
This setting is analogue to anomaly detection: samples that differ significantly from the training data are considered anomalous.
ARGUE concurrently analyses the activation patterns of multiple expert NNs and fuses the insights to one overall anomaly score.
Figuratively speaking, ARGUE moderates between multiple domain experts arguing about the given input sample.
If at least one of these experts has a clear understanding what the input sample means, it is likely normal; if all experts are unsure, it is likely anomalous.
In contrast to the analogy, ARGUE is purely data-driven thus no domain expert knowledge is needed to build the expert NNs.
We visualise our intuition in \Cref{fig:highlevel}.
Our novel architecture automatically decides which expert the input sample belongs to, and judges the impact on the anomaly score.

\subsection{Architecture}
\begin{figure}[tb]
  \begin{center}
    \input{figs/architecture_overview.tikz}
    \caption{
        Architecture of ARGUE in the example of a two-expert setting.
        ARGUE maps the input $\vect{x}$ to an anomaly score $\hat{y}$.
        The shared alarm network analyses the activations of each expert path for anomalous patterns.
        The final anomaly score $\hat{y}$ is the weighted sum of all expert anomaly scores plus the short-cut path, here symbolised by the multiplicative and additive gates. 
    }
    \label{fig:overview}
  \end{center}
\end{figure}
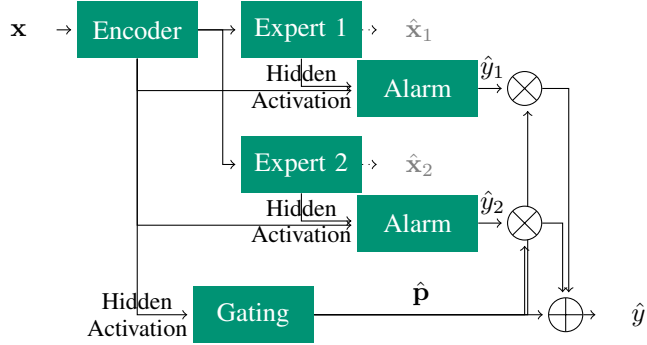

For ARGUE, we combine multiple DNNs to the overall architecture.
At its core, the activations of multiple expert networks are analysed for anomalous behaviour.
An overview of the architecture in the example of a 2-expert system is depicted in \Cref{fig:overview}.
The main components are:
\begin{enumerate}[topsep=0pt,noitemsep]
    \item
    The \emph{encoder} network.
    A DNN reducing the dimensionality of the input.
    It is used as the input to the expert networks and the gating network.
    \item 
    The \emph{expert} networks.
    DNNs trained on reconstructing disjoint parts of the data.
    Combined with the shared encoder network, they work as AEs.
    \item
    The \emph{alarm} network.
    A DNN mapping the activations of the expert paths to an anomaly score.
    There is one shared alarm network.
    \item
    The \emph{gating} network.
    A DNN weighting the importance of each anomaly score.
    It does so by analysing the activations of the encoder network.
\end{enumerate}
The encoder-expert-alarm path is inspired by the target-alarm path in \AAA{}; the gating mechanism is found in ME models.
For this combination to work, we 1)~introduced a shared encoder, 2)~based the gating decision on the activations of the encoder, and 3)~added a short-cut path for easier gradient flow in case of known anomalies.

\ARGUE{} analyses all activation patterns in parallel (using the alarm network) and introduces a way to automatically judge the importance of each anomaly decision (using the gating network).
In the following, we assume a training set $\set{X}$ containing multiple normal classes $\set{X} = \set{X}_1 \cup \set{X}_2 \cup \ldots$, possibly polluted by yet unknown anomalies.
These classes may be known a priori, but can also be approximated by the methods discussed in \Cref{chap:clustering}.

\subsubsection{The Encoder \& Expert Network Form a Multi-headed Autoencoder}
Expert network $j$ learns to reconstruct the input samples $\vect{x} \in \set{X}_j$ given the latent space of the common encoder network.
In other words, we build a multi-headed AE:
\[
    f_\text{expert, j} \circ f_\text{encoder} = f_\text{AE, j}: \vect{x} \mapsto \hat{\vect{x}}_j, \vect{x} \in \set{X}_j.
\]
We train all networks in parallel, thus adapting the weights of the encoder and the expert networks on the respective training samples.
Figuratively, each expert learns the distribution of its normal class bound by the vocabulary of the shared encoder.

\begin{table*}[tb]
\centering
\caption{
Data sets used during evaluation including the applied clustering method and the encoder dimensions.
}\label{tab:data}
\resizebox{\textwidth}{!}{%
\begin{tabular}{c c c c c c c c c}
\textbf{Name} & \textbf{Cl.} & \textbf{Normal}   & \textbf{Anom.} & \textbf{Encoder} & \textbf{$\abs{\set{X}_\text{train, norm}}$} & \textbf{$\abs{\set{X}_\text{test, norm}}$} & \textbf{$\abs{\set{X}_\text{test, anom}}$} & \textbf{A4RGUE} \\
\midrule
CovType \cite{blackard_comparative_1999} & 1\&3 & 1-4 & 5-7 & 90, 75, 60, 45, 25, 15 & \num{4.31e+05} & \num{8.01e+04} & \num{7.07e+03} & 2 \\
EMNIST \cite{cohen_emnist_2017} & 1\&3 & A-M & N-Z  & 16C3-MP2-8C3-MP2-8C3 & \num{5.93e+04} & \num{1.04e+04} & \num{1.04e+04} & 5 \\
FMNIST \cite{xiao_fashion-mnist_2017} & 1\&3 & 0-4 & 5-9 & 16C3-MP2-8C3-MP2-8C3 & \num{2.85e+04} & \num{5.00e+03} & \num{5.00e+03} & 2 \\
MNIST \cite{lecun_gradient-based_1998} & 1\&3 & 0-4 & 5-9 & 16C3-MP2-8C3-MP2-8C3 & \num{2.91e+04} & \num{5.14e+03} & \num{4.86e+03} & 5 \\
\midrule
Census \cite{dua_uci_2017} & 2\&3 & Male, Female & $>50$k & 600, 300, 150, 75, 30, 15 & \num{8.89e+04} & \num{9.36e+04} & \num{6.19e+03} & 4 \\
Darknet \cite{habibi_lashkari_didarknet_2020} & 2\&3 & Prot. 0, 6, 17 & Tor, VPN & 100, 80, 60, 40, 20, 10 & \num{9.46e+04} & \num{1.76e+04} & \num{3.66e+03} & 4 \\
DoH \cite{montazerishatoori_detection_2020} & 2\&3 & Week 1-4 & Malicious & 50, 40, 30, 20, 10, 5 & \num{1.60e+04} & \num{2.94e+03} & \num{3.75e+04} & 5 \\
IDS \cite{sharafaldin_toward_2018} & 2\&3 & Day 1-6 & Bot-In. & 150, 120, 80, 60, 40, 20 & \num{3.46e+06} & \num{6.43e+05} & \num{1.24e+05} & 2 \\
KDD \cite{tavallaee_detailed_2009} & 2\&3 & tcp, udp, icmp & Anomalous & 150, 100, 70, 40, 25, 10 & \num{6.39e+04} & \num{9.71e+03} & \num{1.28e+04} & 5 \\
URL \cite{mamun_detecting_2016} & 2\&3 & Domain 1-2 & Def.-Spam & 100, 80, 60, 40, 20, 10 & \num{6.32e+03} & \num{1.14e+03} & \num{4.37e+03} & 3 \\
\end{tabular}
}
\end{table*}

\subsubsection{The Alarm \& Gating Network Determine the Anomaly Score}
For each expert path, the single shared alarm network analyses the activations and returns an anomaly score $\hat{y}_j$.
Concurrently, the gating network determines the importance of each decision $\hat{\vect{p}}$ based on the activations of the encoder.
\begin{align*}
    f_\text{alarm}:& [\vect{h}_{\text{AE}, j, i}]_i \mapsto \hat{y}_j \in [0,1],\\
    f_\text{gate}:& [\vect{h}_{\text{encoder}, i}]_i \mapsto \hat{\vect{p}}.
\end{align*}
As the gating network is softmax-activated, it returns a probability distribution indicating the most confident expert.
Figuratively, the gating network decides how much to trust each expert's anomaly decision.
The overall output becomes the weighted sum of all scores:
\[
    f_\text{\ARGUE{}}(\vect{x})=\hat{y}=\trans{\hat{\vect{p}}} [\hat{y}_j]_j \in [0,1].
\]

We connected the last element of the gating decision directly to the output.
This tweak allows the gating network to ignore the experts' decision if it already knows the sample to be anomalous, thus creating a short-cut similar to the skip connections in ResNet \cite{he_deep_2016}.
If the gating network identifies the sample as unknown, the overall decision is shifted to anomalous by the auxiliary path;
else the decision is handed to the expert networks.
We evaluate the influence of the short-cut path in our ablation study.

\subsection{Training Objectives}
We combine all components to the overall architecture of ARGUE.
First, we pretrain the expert paths of the given normal classes.
Next, the hidden activations are analysed by the gating and alarm network.
The gating network is trained to return $1_j$ when the input belongs to class $\vect{x} \in \set{X}_j$, where $1_j$ denotes a 1-Hot vector with~1 at position $j$ and~0 elsewhere.
Let this function be $\vect{p}(\vect{x})$.

AD is characterised by its inherent class imbalance, where anomalous training samples are rare.
As done in \AAA{}, a Gaussian prior generates noise samples $\tilde{\vect{x}} \sim \set{N}( 0.5, 1 )$ used as artificial counterexamples during training of the AD-related components.
All normal training samples are scaled to $\vect{x} \in [0,1]$, thus $\tilde{\vect{x}}$ is likely outside of the normal input ranges.
Whenever such a noise sample is at the input, the training label becomes $y=1$ with the gating target $\vect{p}_\text{anom}=[0, \ldots, 0, 1]$, i.e. prioritising the short-cut path.
Thanks to the noise prior, our AD problem is reduced to two classifications: the binary anomaly score $\hat{y}$ and the multi-class gating decision $\hat{\vect{p}}$.
We use the binary (BX) and categorical (CX) cross-entropy as loss functions:
\begin{equation*}
        \argmin_{\theta_\text{alarm}} \Expected [
\set{L}_\text{BX} \left( y, f_\text{\ARGUE{}} \left( \vect{x} \right) \right)
+ \set{L}_\text{BX} \left( 1, f_\text{\ARGUE{}} \left( \tilde{\vect{x}} \right) \right)
]
\end{equation*}
\begin{equation*}
\argmin_{\theta_\text{gate}} \Expected [
\left(
\set{L}_\text{CX} \left( \vect{p}(\vect{x}), f_\text{gate} \left( \vect{x} \right) \right)
+ \set{L}_\text{CX} \left( \vect{p}_\text{anom}, f_\text{gate} \left( \tilde{\vect{x}} \right) \right)
\right],
\end{equation*}
where $(\vect{x},y) \sim P_\set{D}, \tilde{\vect{x}} \sim \set{N}( 0.5, 1 )$.
For unsupervised settings, where no anomaly-related labels are available, we fix $y \equiv 0$ and assume that all input data is normal even when polluted by noise.

\section{Experimental Setup}

\subsection{Data Sets}
\label{chap:clustering}
We chose ten challenging data sets to evaluate the performance of ARGUE.
In the following, we discuss three ways to distribute the data among our expert NNs.
\Cref{tab:data} gives an overview about the clustering method used for each data set denoted by ``Cl.''.

\begin{enumerate}[topsep=0pt,noitemsep]
    \item \emph{By Class}:
    under ideal conditions, suitable clusters are known a priori.
    Each expert NN focuses on one class.
\end{enumerate}

We acknowledge that an ideal clustering is usually only available in OOD problems.
To expand \ARGUE{} to AD, we apply two strategies to estimate suitable clusters based on the data itself, i.e. in an unsupervised/self-supervised way:

\begin{enumerate}[topsep=0pt,noitemsep]
    \setcounter{enumi}{1}
    \item \emph{By Attribute}:
    the data itself may contain attributes for clusters.
    We evaluated several attributes, e.g. the protocol types of network packets.
    \item \emph{By Algorithms}:
    when none of the above is applicable, suitable clustering algorithms may be used.
\end{enumerate}

The chosen data sets cover all three scenarios.
Whereas the upper part in \Cref{tab:data} contains common classification data sets with \emph{multiple known clusters}, the lower part summarises AD data sets, which comprise \emph{one normal class}.
In the lower part of \Cref{tab:data}, we show how we distributed the notion of normal when applying scenario~2.
We believe that most real-world data sets contain some intuitive way to split the data.
For example, the considered network data sets were recorded over several days --
it may be favourable to take each day as separate notion of normal.
Naturally, automatic clustering, i.e. scenario~3, is applicable to all data sets.

\paragraph*{Automatic clustering}
We motivate scenario~3 by applying an unsupervised adversarial autoencoder (AAE) \cite{makhzani_adversarial_2016}.
AAEs are AEs, where the latent space is conditioned on a Gaussian distribution.
By introducing a separate discriminator, they can also be used for clustering.
We would love to see future work by the clustering community on algorithms covering AD as use case.
During our evaluation, we call the adversarial autoencoder assisted version A4RGUE.
The selected number of clusters are shown in \Cref{tab:data}, which we based on the validation scores.

\subsection{Baseline Methods}
ARGUE applies to several different AD scenarios.
We chose ten baseline methods of all related categories.
For a fair comparison, we used the very same configuration for the baseline methods as done in ARGUE if applicable.
If the same data set was evaluated by the original authors, we used their configuration instead.
Our unsupervised baseline methods are the following:
\ARGUE{} is based on AEs, thus we use them as simple DL-based baseline.
The reconstruction error is used as anomaly score.
For the advanced DL-based AD methods, we chose Deep-SVDD \cite{ruff_deep_2018}, a one-class classifier, f-AnoGAN \cite{schlegl_f-anogan_2019} \& GANomaly \cite{akcay_ganomaly_2019}, two GAN-based approaches, DAGMM \cite{zong_deep_2018}, combining AEs and GMMs, and MEx-CVAEC \cite{yu_mixture_2021}, a ME-based AD method.
For a complementary view, we include the outlier detection method REPEN \cite{pang_learning_2018}.
In semi-supervised scenarios, we compare to three popular baselines:
DeepSAD \cite{ruff_deep_2020}, Deviation Networks (DevNet) \cite{pang_deep_2019} and \AAA{} \cite{sperl_activation_2021}.

\section{Evaluation}
In the following, we discuss ARGUE's performance compared to other state-of-the-art AD methods.
We closely follow the best-practices introduced by Hendrycks~\&~Gimpel \cite{hendrycks_baseline_2017} and measure the performance as area under the ROC curve (AUC) as well as average precision (AP).
The AUC gives the trade-off between the true and false positive rate.
Both metrics work independent of a detection threshold.
In both cases, $1$ is the highest score.
Additionally, we report the p-value of the Wilcoxon signed-rank test \cite{wilcoxon_individual_1992} to show the significance of our results.
It evaluates the hypothesis if the measurements were derived from the same distribution.

\subsection{Multiple Classes of Normal}
\label{chap:motivation}
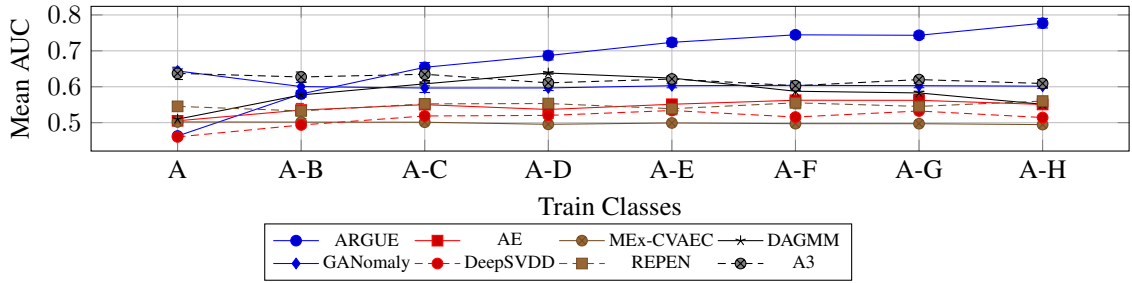
\begin{figure}[tb]
  \begin{center}
    \input{figs/motivation_plot.tikz}
    \caption{
        Unsupervised AD performance on EMNIST measured as AUC after 5 runs.
        We defined the letters N-Z to be anomalous.
        For each experiment, we increased the number of normal classes within the training set and added the respective expert.
        \ARGUE{} was the only AD method in our evaluation profiting from the additional knowledge.
    }
    \label{fig:motivation}
  \end{center}
\end{figure}

We begin our evaluation by motivating \ARGUE{}'s multi-expert architecture.
Our initial assumption was that anomalies are easier to detect when we learn the different notions of normal separately using individual experts.
In \Cref{fig:motivation}, we increased the amount of training classes in the EMNIST data set and test against the very same anomalies, here: the letters N-Z.
Usual AD methods need to adapt to all notions of normal within one model, whereas ARGUE distributes the knowledge across multiple expert paths.

For one class, ARGUE scored an AUC of 46\%, which improved to 77\% for eight normal classes.
ARGUE surpassed the baselines at the point where three normal classes were present in the training set.
The baseline methods remained approximately at their initial performance.
Although each new normal class increases the amount of training data, the baseline methods struggled to profit from it.
The rising complexity of modelling multiple types of normal seemed to thwart any AD performance gains.
In contrast, ARGUE leveraged the prior knowledge of the multiple notions of normal.

We feel reassured that ARGUE tackles an important problem in AD:
real-world data sets may contain multiple processes, where the data originates from, e.g. multiple computers in a network.
Current AD methods work on these data sets, yet do not profit from prior knowledge about the distribution of the normal classes.
In the following experiments, we apply \ARGUE{} under diverse scenarios, gradually decreasing the amount of prior knowledge available.

\begin{table*}[tb]
\centering
\setlength{\tabcolsep}{3pt}
\caption{
Semi-supervised test results: mean AUC \& AP incl. the standard deviation after 5 runs.
}\label{tab:semi_results}
\resizebox{\textwidth}{!}{%
\begin{tabular}{>{\color{gray}}c >{\color{gray}}c >{\color{gray}}c >{\color{gray}}c >{\color{gray}}c >{\color{gray}}c | >{\color{gray}}c >{\color{gray}}c >{\color{gray}}c >{\color{gray}}c | >{\color{gray}}c >{\color{gray}}c >{\color{gray}}c >{\color{gray}}c  >{\color{gray}}c >{\color{gray}}c }
    &  & \multicolumn{4}{>{\color{gray}}c}{Ours} &                            \multicolumn{4}{>{\color{gray}}c}{Unsupervised Baselines} &                            \multicolumn{6}{>{\color{gray}}c}{Semi-supervised Baselines} \\

    & Method & \multicolumn{2}{c}{ARGUE} & \multicolumn{2}{c}{A4RGUE} & \multicolumn{2}{c}{AE} & \multicolumn{2}{c}{MEx-CVAEC} & \multicolumn{2}{c}{DeepSAD} & \multicolumn{2}{c}{DevNet} & \multicolumn{2}{c}{A3} \\
    & Metric &                                              AUC &                                               AP &                                 AUC &                                               AP &                                 AUC &                                  AP &                                              AUC &                                               AP &                                 AUC &                                  AP &                                              AUC &                                               AP &                                              AUC &                                               AP \\
\midrule
&  CovType &  \color{black}$.83 \scriptscriptstyle \pm .02$ &  \color{black}$.46 \scriptscriptstyle \pm .03$ &  $.80 \scriptscriptstyle \pm .03$ &               $.40 \scriptscriptstyle \pm .03$ &  $.62 \scriptscriptstyle \pm .02$ &  $.16 \scriptscriptstyle \pm .01$ &               $.75 \scriptscriptstyle \pm .01$ &               $.21 \scriptscriptstyle \pm .02$ &  $.65 \scriptscriptstyle \pm .05$ &  $.21 \scriptscriptstyle \pm .02$ &  \color{black}$.83 \scriptscriptstyle \pm .00$ &               $.33 \scriptscriptstyle \pm .03$ &               $.80 \scriptscriptstyle \pm .01$ &               $.42 \scriptscriptstyle \pm .01$ \\
    &  EMNIST &  \color{black}$.92 \scriptscriptstyle \pm .01$ &  \color{black}$.92 \scriptscriptstyle \pm .00$ &  $.84 \scriptscriptstyle \pm .01$ &               $.85 \scriptscriptstyle \pm .01$ &  $.54 \scriptscriptstyle \pm .01$ &  $.53 \scriptscriptstyle \pm .00$ &               $.50 \scriptscriptstyle \pm .01$ &               $.50 \scriptscriptstyle \pm .00$ &  $.56 \scriptscriptstyle \pm .01$ &  $.57 \scriptscriptstyle \pm .01$ &               $.81 \scriptscriptstyle \pm .01$ &               $.82 \scriptscriptstyle \pm .01$ &               $.85 \scriptscriptstyle \pm .01$ &               $.87 \scriptscriptstyle \pm .01$ \\
    &  FMNIST &  \color{black}$.95 \scriptscriptstyle \pm .01$ &  \color{black}$.96 \scriptscriptstyle \pm .01$ &  $.93 \scriptscriptstyle \pm .01$ &               $.95 \scriptscriptstyle \pm .01$ &  $.83 \scriptscriptstyle \pm .01$ &  $.81 \scriptscriptstyle \pm .01$ &               $.74 \scriptscriptstyle \pm .01$ &               $.74 \scriptscriptstyle \pm .01$ &  $.74 \scriptscriptstyle \pm .03$ &  $.76 \scriptscriptstyle \pm .03$ &               $.94 \scriptscriptstyle \pm .01$ &               $.95 \scriptscriptstyle \pm .01$ &  \color{black}$.95 \scriptscriptstyle \pm .01$ &  \color{black}$.96 \scriptscriptstyle \pm .01$ \\
    &  MNIST &  \color{black}$.99 \scriptscriptstyle \pm .00$ &  \color{black}$.99 \scriptscriptstyle \pm .00$ &  $.96 \scriptscriptstyle \pm .01$ &               $.97 \scriptscriptstyle \pm .01$ &  $.76 \scriptscriptstyle \pm .01$ &  $.73 \scriptscriptstyle \pm .01$ &               $.62 \scriptscriptstyle \pm .01$ &               $.58 \scriptscriptstyle \pm .01$ &  $.61 \scriptscriptstyle \pm .03$ &  $.58 \scriptscriptstyle \pm .02$ &               $.94 \scriptscriptstyle \pm .01$ &               $.93 \scriptscriptstyle \pm .01$ &               $.97 \scriptscriptstyle \pm .01$ &               $.98 \scriptscriptstyle \pm .01$ \\

\midrule

    & Census &               $.80 \scriptscriptstyle \pm .03$ &               $.26 \scriptscriptstyle \pm .04$ &  $.80 \scriptscriptstyle \pm .02$ &               $.27 \scriptscriptstyle \pm .03$ &  $.68 \scriptscriptstyle \pm .01$ &  $.09 \scriptscriptstyle \pm .00$ &               $.64 \scriptscriptstyle \pm .01$ &               $.08 \scriptscriptstyle \pm .00$ &  $.72 \scriptscriptstyle \pm .03$ &  $.22 \scriptscriptstyle \pm .05$ &  \color{black}$.90 \scriptscriptstyle \pm .00$ &  \color{black}$.42 \scriptscriptstyle \pm .02$ &               $.84 \scriptscriptstyle \pm .01$ &               $.27 \scriptscriptstyle \pm .04$ \\
    & Darknet &               $.89 \scriptscriptstyle \pm .01$ &  \color{black}$.71 \scriptscriptstyle \pm .01$ &  $.86 \scriptscriptstyle \pm .01$ &               $.68 \scriptscriptstyle \pm .01$ &  $.61 \scriptscriptstyle \pm .03$ &  $.28 \scriptscriptstyle \pm .01$ &               $.49 \scriptscriptstyle \pm .01$ &               $.19 \scriptscriptstyle \pm .00$ &  $.69 \scriptscriptstyle \pm .12$ &  $.38 \scriptscriptstyle \pm .14$ &  \color{black}$.90 \scriptscriptstyle \pm .02$ &               $.69 \scriptscriptstyle \pm .02$ &               $.87 \scriptscriptstyle \pm .01$ &               $.69 \scriptscriptstyle \pm .01$ \\
    & DoH &  \color{black}$1.00 \scriptscriptstyle \pm .00$ &  \color{black}$1.00 \scriptscriptstyle \pm .00$ &  $.99 \scriptscriptstyle \pm .01$ &  \color{black}$1.00 \scriptscriptstyle \pm .00$ &  $.89 \scriptscriptstyle \pm .02$ &  $.99 \scriptscriptstyle \pm .00$ &               $.78 \scriptscriptstyle \pm .08$ &               $.97 \scriptscriptstyle \pm .01$ &  $.91 \scriptscriptstyle \pm .14$ &  $.99 \scriptscriptstyle \pm .01$ &               $.90 \scriptscriptstyle \pm .03$ &               $.98 \scriptscriptstyle \pm .01$ &  \color{black}$1.00 \scriptscriptstyle \pm .00$ &  \color{black}$1.00 \scriptscriptstyle \pm .00$ \\
    & IDS &  \color{black}$.98 \scriptscriptstyle \pm .00$ &  \color{black}$.94 \scriptscriptstyle \pm .00$ &  $.85 \scriptscriptstyle \pm .09$ &               $.80 \scriptscriptstyle \pm .12$ &  $.90 \scriptscriptstyle \pm .01$ &  $.72 \scriptscriptstyle \pm .03$ &               $.66 \scriptscriptstyle \pm .13$ &               $.27 \scriptscriptstyle \pm .08$ &  $.85 \scriptscriptstyle \pm .04$ &  $.75 \scriptscriptstyle \pm .05$ &               $.88 \scriptscriptstyle \pm .02$ &               $.67 \scriptscriptstyle \pm .10$ &               $.96 \scriptscriptstyle \pm .01$ &               $.92 \scriptscriptstyle \pm .01$ \\
    & KDD &               $.86 \scriptscriptstyle \pm .04$ &               $.89 \scriptscriptstyle \pm .04$ &  $.84 \scriptscriptstyle \pm .05$ &               $.90 \scriptscriptstyle \pm .03$ &  $.93 \scriptscriptstyle \pm .01$ &  $.93 \scriptscriptstyle \pm .01$ &  \color{black}$.95 \scriptscriptstyle \pm .01$ &  \color{black}$.95 \scriptscriptstyle \pm .01$ &  $.80 \scriptscriptstyle \pm .19$ &  $.87 \scriptscriptstyle \pm .12$ &               $.92 \scriptscriptstyle \pm .01$ &               $.93 \scriptscriptstyle \pm .01$ &               $.84 \scriptscriptstyle \pm .04$ &               $.89 \scriptscriptstyle \pm .02$ \\
    & URL &  \color{black}$.96 \scriptscriptstyle \pm .01$ &  \color{black}$.99 \scriptscriptstyle \pm .00$ &  $.94 \scriptscriptstyle \pm .01$ &  \color{black}$.99 \scriptscriptstyle \pm .00$ &  $.92 \scriptscriptstyle \pm .01$ &  $.98 \scriptscriptstyle \pm .00$ &               $.87 \scriptscriptstyle \pm .01$ &               $.96 \scriptscriptstyle \pm .00$ &  $.79 \scriptscriptstyle \pm .13$ &  $.95 \scriptscriptstyle \pm .03$ &               $.94 \scriptscriptstyle \pm .01$ &               $.98 \scriptscriptstyle \pm .00$ &               $.95 \scriptscriptstyle \pm .01$ &  \color{black}$.99 \scriptscriptstyle \pm .00$ \\
    & mean &                              \color{black}$.92$ &                              \color{black}$.81$ &                              $.88$ &                                           $.78$ &                              $.77$ &                              $.62$ &                                           $.70$ &                                           $.54$ &                              $.73$ &                              $.63$ &                                           $.90$ &                                           $.77$ &                                           $.90$ &                                           $.80$ \\
    & p-val &                                                - &                                                - &                              $.00$ &                                           $.04$ &                              $.01$ &                              $.01$ &                                           $.01$ &                                           $.01$ &                              $.00$ &                              $.00$ &                                           $.38$ &                                           $.19$ &                                           $.11$ &                                           $.19$ \\

\end{tabular}
}
\end{table*}

\subsection{Semi-supervised AD}
\label{chap:semi}
In real-world scenarios a few known anomalous examples may already be known.
Semi-supervised methods leverage this type of prior knowledge to boost the AD performance.
In our evaluation, we limited the amount of known anomalies to 100, randomly sampled from all available ones, i.e. much less than the available normal samples.
We show the results in \Cref{tab:semi_results}.

All semi-supervised methods performed on a high level, yet we are happy to report that \ARGUE{} is the best performing one with a mean AUC of 92\%.
The performance advantage is the largest for \emph{known clusters} summarised in the upper half of the table.
Here, \ARGUE{} is \percincrease{85}{91}better than its follow-up \AAA{} on EMNIST, which contained 13 different normal classes.
In real-world scenarios, ideal clusters may not be available, thus we expanded our experiments to \emph{estimated clusters} summarised in the lower half of the table.
Here, we split the data based on features of the input data according to \Cref{tab:data}, i.e. information available to all AD methods.
On all data sets except FMNIST and Census \ARGUE{} performed better than \AAA{}, which we are improving on.

Finally, in the column marked by A4RGUE we show the performance after \emph{automatic clustering} using AAEs.
The results became more volatile, presumably due to the different clustering results during the runs.
As discussed by Ye~et~al. \cite{ye_understanding_2021}, some known anomalies may even reduce the AD performance, when being too close to normal ones.
Although A4RGUE reached \ARGUE{}'s performance on e.g. Census and DoH, the mean AUC decreased slightly to 88\%.
For completeness, we also showed the performance of the most related unsupervised baseline methods.
Although not reaching the same mean performance, we saw competitive results on some data sets, e.g. FMNIST and KDD.
Concluding our semi-supervised experiments, \ARGUE{} profited from both types of prior knowledge: anomaly labels and clusters in the normal data.
Supervised and self-supervised clusters resulted in stronger performance than using a simple clustering algorithm.

\begin{table*}[tb]
\centering
\setlength{\tabcolsep}{1pt}
\caption{
Unsuperv. test results: mean AUC \& AP incl. the standard deviation after 5 runs.
Not all methods scaled to IDS.
}\label{tab:unsu_results}

\resizebox{\textwidth}{!}{%

\begin{tabular}{>{\color{gray}}c >{\color{gray}}c >{\color{gray}}c >{\color{gray}}c >{\color{gray}}c >{\color{gray}}c | >{\color{gray}}c >{\color{gray}}c >{\color{gray}}c >{\color{gray}}c >{\color{gray}}c >{\color{gray}}c >{\color{gray}}c >{\color{gray}}c >{\color{gray}}c >{\color{gray}}c >{\color{gray}}c >{\color{gray}}c >{\color{gray}}c >{\color{gray}}c | >{\color{gray}}c >{\color{gray}}c }

    &  & \multicolumn{4}{>{\color{gray}}c}{Ours} &                            \multicolumn{14}{>{\color{gray}}c}{Unsupervised Baselines} &                            \multicolumn{2}{>{\color{gray}}c}{Ablation} \\

     & Method & \multicolumn{2}{c}{ARGUE} & \multicolumn{2}{c}{A4RGUE} & \multicolumn{2}{c}{AE} & \multicolumn{2}{c}{MEx-CVAEC} & \multicolumn{2}{c}{DAGMM} & \multicolumn{2}{c}{GANomaly} & \multicolumn{2}{c}{fAnoGAN} & \multicolumn{2}{c}{DeepSVDD} & \multicolumn{2}{c}{REPEN} & \multicolumn{2}{c}{A3} \\
     & Metric &                                              AUC &                                               AP &                                 AUC &                                  AP &                                              AUC &                                               AP &                                              AUC &                                               AP &                                 AUC &                                  AP &                                 AUC &                                  AP &                                 AUC &                                  AP &                                 AUC &                                  AP &                                 AUC &                                  AP &                                 AUC &                                  AP \\
\midrule
 &  CovType &  \color{black}$.94 \scriptscriptstyle \pm .00$ &  \color{black}$.67 \scriptscriptstyle \pm .02$ &  $.62 \scriptscriptstyle \pm .08$ &  $.15 \scriptscriptstyle \pm .06$ &               $.59 \scriptscriptstyle \pm .02$ &               $.14 \scriptscriptstyle \pm .01$ &               $.75 \scriptscriptstyle \pm .02$ &               $.19 \scriptscriptstyle \pm .02$ &  $.71 \scriptscriptstyle \pm .04$ &  $.19 \scriptscriptstyle \pm .03$ &  $.64 \scriptscriptstyle \pm .05$ &  $.13 \scriptscriptstyle \pm .03$ &  $.60 \scriptscriptstyle \pm .06$ &  $.12 \scriptscriptstyle \pm .02$ &  $.57 \scriptscriptstyle \pm .04$ &  $.12 \scriptscriptstyle \pm .02$ &  $.72 \scriptscriptstyle \pm .02$ &  $.16 \scriptscriptstyle \pm .01$ &  $.55 \scriptscriptstyle \pm .10$ &  $.10 \scriptscriptstyle \pm .03$ \\
     &  EMNIST &  \color{black}$.90 \scriptscriptstyle \pm .02$ &  \color{black}$.89 \scriptscriptstyle \pm .02$ &  $.58 \scriptscriptstyle \pm .02$ &  $.57 \scriptscriptstyle \pm .02$ &               $.53 \scriptscriptstyle \pm .01$ &               $.53 \scriptscriptstyle \pm .00$ &               $.49 \scriptscriptstyle \pm .00$ &               $.49 \scriptscriptstyle \pm .00$ &  $.53 \scriptscriptstyle \pm .01$ &  $.54 \scriptscriptstyle \pm .01$ &  $.62 \scriptscriptstyle \pm .02$ &  $.58 \scriptscriptstyle \pm .01$ &  $.53 \scriptscriptstyle \pm .04$ &  $.51 \scriptscriptstyle \pm .02$ &  $.54 \scriptscriptstyle \pm .01$ &  $.55 \scriptscriptstyle \pm .01$ &  $.52 \scriptscriptstyle \pm .03$ &  $.52 \scriptscriptstyle \pm .01$ &  $.60 \scriptscriptstyle \pm .03$ &  $.57 \scriptscriptstyle \pm .02$ \\
     &  FMNIST &  \color{black}$.87 \scriptscriptstyle \pm .02$ &  \color{black}$.87 \scriptscriptstyle \pm .02$ &  $.68 \scriptscriptstyle \pm .08$ &  $.60 \scriptscriptstyle \pm .06$ &               $.81 \scriptscriptstyle \pm .01$ &               $.79 \scriptscriptstyle \pm .01$ &               $.70 \scriptscriptstyle \pm .01$ &               $.69 \scriptscriptstyle \pm .01$ &  $.82 \scriptscriptstyle \pm .03$ &  $.76 \scriptscriptstyle \pm .03$ &  $.82 \scriptscriptstyle \pm .03$ &  $.81 \scriptscriptstyle \pm .04$ &  $.72 \scriptscriptstyle \pm .13$ &  $.71 \scriptscriptstyle \pm .15$ &  $.68 \scriptscriptstyle \pm .01$ &  $.70 \scriptscriptstyle \pm .01$ &  $.70 \scriptscriptstyle \pm .14$ &  $.64 \scriptscriptstyle \pm .11$ &  $.81 \scriptscriptstyle \pm .07$ &  $.75 \scriptscriptstyle \pm .08$ \\
     &  MNIST &  \color{black}$.98 \scriptscriptstyle \pm .00$ &  \color{black}$.98 \scriptscriptstyle \pm .00$ &  $.69 \scriptscriptstyle \pm .09$ &  $.65 \scriptscriptstyle \pm .08$ &               $.75 \scriptscriptstyle \pm .01$ &               $.72 \scriptscriptstyle \pm .01$ &               $.60 \scriptscriptstyle \pm .01$ &               $.56 \scriptscriptstyle \pm .01$ &  $.72 \scriptscriptstyle \pm .03$ &  $.72 \scriptscriptstyle \pm .02$ &  $.77 \scriptscriptstyle \pm .04$ &  $.73 \scriptscriptstyle \pm .04$ &  $.52 \scriptscriptstyle \pm .07$ &  $.50 \scriptscriptstyle \pm .06$ &  $.57 \scriptscriptstyle \pm .01$ &  $.55 \scriptscriptstyle \pm .01$ &  $.49 \scriptscriptstyle \pm .01$ &  $.45 \scriptscriptstyle \pm .01$ &  $.45 \scriptscriptstyle \pm .01$ &  $.43 \scriptscriptstyle \pm .01$ \\

\midrule

     & Census &  \color{black}$.85 \scriptscriptstyle \pm .02$ &  \color{black}$.34 \scriptscriptstyle \pm .02$ &  $.64 \scriptscriptstyle \pm .09$ &  $.09 \scriptscriptstyle \pm .02$ &               $.67 \scriptscriptstyle \pm .01$ &               $.09 \scriptscriptstyle \pm .00$ &               $.63 \scriptscriptstyle \pm .01$ &               $.08 \scriptscriptstyle \pm .00$ &  $.39 \scriptscriptstyle \pm .08$ &  $.05 \scriptscriptstyle \pm .01$ &  $.52 \scriptscriptstyle \pm .10$ &  $.07 \scriptscriptstyle \pm .01$ &  $.56 \scriptscriptstyle \pm .07$ &  $.08 \scriptscriptstyle \pm .02$ &  $.58 \scriptscriptstyle \pm .02$ &  $.09 \scriptscriptstyle \pm .00$ &  $.65 \scriptscriptstyle \pm .03$ &  $.08 \scriptscriptstyle \pm .01$ &  $.61 \scriptscriptstyle \pm .04$ &  $.08 \scriptscriptstyle \pm .01$ \\
     & Darknet &  \color{black}$.95 \scriptscriptstyle \pm .00$ &  \color{black}$.77 \scriptscriptstyle \pm .01$ &  $.65 \scriptscriptstyle \pm .08$ &  $.25 \scriptscriptstyle \pm .07$ &               $.58 \scriptscriptstyle \pm .03$ &               $.26 \scriptscriptstyle \pm .01$ &               $.51 \scriptscriptstyle \pm .05$ &               $.19 \scriptscriptstyle \pm .01$ &  $.50 \scriptscriptstyle \pm .05$ &  $.18 \scriptscriptstyle \pm .01$ &  $.63 \scriptscriptstyle \pm .10$ &  $.26 \scriptscriptstyle \pm .06$ &  $.52 \scriptscriptstyle \pm .13$ &  $.18 \scriptscriptstyle \pm .03$ &  $.51 \scriptscriptstyle \pm .17$ &  $.21 \scriptscriptstyle \pm .06$ &  $.34 \scriptscriptstyle \pm .02$ &  $.16 \scriptscriptstyle \pm .03$ &  $.44 \scriptscriptstyle \pm .09$ &  $.17 \scriptscriptstyle \pm .02$ \\
     & DoH &  \color{black}$.95 \scriptscriptstyle \pm .02$ &  \color{black}$.99 \scriptscriptstyle \pm .00$ &  $.68 \scriptscriptstyle \pm .05$ &  $.96 \scriptscriptstyle \pm .01$ &               $.89 \scriptscriptstyle \pm .04$ &  \color{black}$.99 \scriptscriptstyle \pm .01$ &               $.77 \scriptscriptstyle \pm .09$ &               $.97 \scriptscriptstyle \pm .01$ &  $.65 \scriptscriptstyle \pm .07$ &  $.95 \scriptscriptstyle \pm .01$ &  $.65 \scriptscriptstyle \pm .09$ &  $.95 \scriptscriptstyle \pm .01$ &  $.54 \scriptscriptstyle \pm .14$ &  $.92 \scriptscriptstyle \pm .03$ &  $.64 \scriptscriptstyle \pm .05$ &  $.96 \scriptscriptstyle \pm .01$ &  $.14 \scriptscriptstyle \pm .03$ &  $.86 \scriptscriptstyle \pm .01$ &  $.82 \scriptscriptstyle \pm .07$ &  $.98 \scriptscriptstyle \pm .01$ \\
     & IDS &  \color{black}$.96 \scriptscriptstyle \pm .02$ &  \color{black}$.92 \scriptscriptstyle \pm .02$ &  $.67 \scriptscriptstyle \pm .07$ &  $.33 \scriptscriptstyle \pm .10$ &               $.74 \scriptscriptstyle \pm .06$ &               $.29 \scriptscriptstyle \pm .05$ &               $.54 \scriptscriptstyle \pm .11$ &               $.20 \scriptscriptstyle \pm .07$ &    -- &    -- &  $.65 \scriptscriptstyle \pm .06$ &  $.24 \scriptscriptstyle \pm .04$ &  $.46 \scriptscriptstyle \pm .17$ &  $.20 \scriptscriptstyle \pm .14$ &  $.51 \scriptscriptstyle \pm .14$ &  $.21 \scriptscriptstyle \pm .13$ &    -- &    -- &  $.52 \scriptscriptstyle \pm .05$ &  $.19 \scriptscriptstyle \pm .05$ \\
     & KDD &               $.91 \scriptscriptstyle \pm .01$ &               $.87 \scriptscriptstyle \pm .04$ &  $.83 \scriptscriptstyle \pm .10$ &  $.83 \scriptscriptstyle \pm .08$ &               $.91 \scriptscriptstyle \pm .01$ &               $.90 \scriptscriptstyle \pm .02$ &  \color{black}$.94 \scriptscriptstyle \pm .01$ &  \color{black}$.94 \scriptscriptstyle \pm .02$ &  $.88 \scriptscriptstyle \pm .02$ &  $.89 \scriptscriptstyle \pm .03$ &  $.86 \scriptscriptstyle \pm .07$ &  $.87 \scriptscriptstyle \pm .06$ &  $.62 \scriptscriptstyle \pm .25$ &  $.73 \scriptscriptstyle \pm .17$ &  $.81 \scriptscriptstyle \pm .14$ &  $.86 \scriptscriptstyle \pm .10$ &  $.85 \scriptscriptstyle \pm .03$ &  $.86 \scriptscriptstyle \pm .03$ &  $.58 \scriptscriptstyle \pm .07$ &  $.59 \scriptscriptstyle \pm .07$ \\
     & URL &               $.92 \scriptscriptstyle \pm .02$ &  \color{black}$.98 \scriptscriptstyle \pm .01$ &  $.64 \scriptscriptstyle \pm .07$ &  $.86 \scriptscriptstyle \pm .03$ &  \color{black}$.93 \scriptscriptstyle \pm .01$ &  \color{black}$.98 \scriptscriptstyle \pm .00$ &               $.88 \scriptscriptstyle \pm .02$ &               $.96 \scriptscriptstyle \pm .01$ &  $.75 \scriptscriptstyle \pm .02$ &  $.91 \scriptscriptstyle \pm .01$ &  $.79 \scriptscriptstyle \pm .05$ &  $.93 \scriptscriptstyle \pm .02$ &  $.81 \scriptscriptstyle \pm .02$ &  $.93 \scriptscriptstyle \pm .01$ &  $.74 \scriptscriptstyle \pm .02$ &  $.92 \scriptscriptstyle \pm .01$ &  $.73 \scriptscriptstyle \pm .04$ &  $.90 \scriptscriptstyle \pm .02$ &  $.84 \scriptscriptstyle \pm .02$ &  $.95 \scriptscriptstyle \pm .01$ \\
     & mean &                              \color{black}$.92$ &                              \color{black}$.83$ &                              $.67$ &                              $.53$ &                                           $.74$ &                                           $.57$ &                                           $.68$ &                                           $.53$ &                              $.66$ &                              $.58$ &                              $.69$ &                              $.56$ &                              $.59$ &                              $.49$ &                              $.62$ &                              $.52$ &                              $.57$ &                              $.51$ &                              $.62$ &                              $.48$ \\
     & p-val &                                                -- &                                                -- &                              $.00$ &                              $.00$ &                                           $.01$ &                                           $.01$ &                                           $.00$ &                                           $.01$ &                              $.08$ &                              $.11$ &                              $.00$ &                              $.00$ &                              $.00$ &                              $.00$ &                              $.00$ &                              $.00$ &                              $.08$ &                              $.08$ &                              $.00$ &                              $.00$ \\

\end{tabular}

}

\end{table*}

\subsection{Unsupervised AD}
\label{chap:unsu}
Unsupervised scenarios are regarded as the core application for AD.
For real-world scenarios, it is usually infeasible to guarantee that all training samples are normal, which we simulated by polluting 1\% of the data by anomalies.
In our third experiment, we removed all known anomalies from the training data.
The results are summarised in \Cref{tab:unsu_results}.
Here, we measured the influence of the second type of prior knowledge in \ARGUE{}: possible clusters in the training data.

We are happy to report that \ARGUE{} scored the best with a mean AUC of 92\%.
This is a \percincrease{74}{92}increase to the follow-up method, the AE.
For the \emph{known clusters}, \ARGUE{} reached its semi-supervised performance, and even surpassed it on CovType.
Even more interesting, we saw the same performance increase on the data sets with \emph{estimated clusters}.
Unlike OOD detection, anomalies will also belong to one of these clusters.
\ARGUE{} scored the best on all data sets except URL and KDD.
On Census, where we split between male and female citizens, \ARGUE{} was \percincrease{67}{85}better than its follow-up, the AE; on Darknet, where we used protocol types e.g. TCP, \percincrease{63}{95}better than GANomaly.
We like to stress that the attributes of the estimated clusters are in the input data, thus available to all AD methods.
Again, we expanded our research to \emph{automatic clustering}.
A4RGUE's mean performance was on a similar level as the unsupervised baseline methods, but only beat them on Darknet.
The increase in variance for A4RGUE may be a sign that the AAE's cluster predictions changed during the runs -- some splits resulted in better AD performance.
Although both the self-supervised and unsupervised approach work on unlabelled data regarding the clusters, the first one performed significantly better.
We hope to spark interest in the clustering community to bridge the gap between the two approaches.

\ARGUE{}'s performance is strong evidence that \emph{it may be as favourable to estimate normal clusters as manually finding anomalies}.
Real-world data sets often contain knowledge about possible clusters ``for free'', e.g. temporal or spatial information.
We used several possible attributes in our experiments as seen in \cref{tab:data}.
On Census, Darknet and KDD, \ARGUE{} performed better using only these estimated clusters instead of known anomalies.
In some scenarios, it may be more feasible to introduce some assumptions on the distribution of the normal data instead of searching for anomalies among it.
\ARGUE{} allows to use both types of prior knowledge.

\subsection{Ablation Study}\label{sec:ablation}

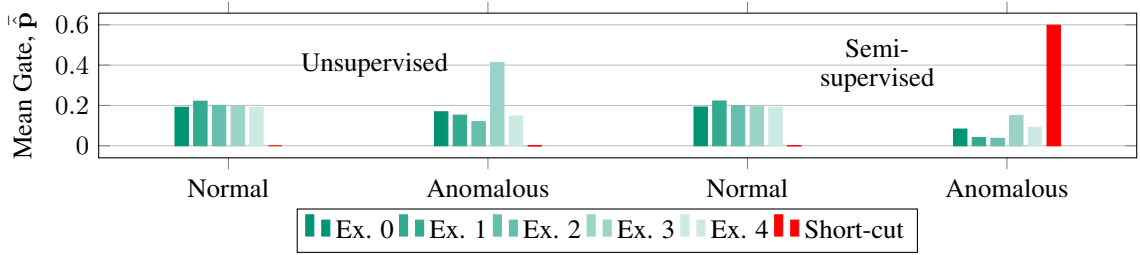
\begin{figure}[tb]
  \begin{center}
    \input{figs/ablation_gate.tikz}
    \caption{
    Mean outputs of the gating network across all test samples on MNIST.
    Digits 0-4 were defined to be normal, resulting in 5 expert NNs plus the short-cut path.
    }
    \label{fig:ablation}
  \end{center}
\end{figure}

\paragraph{Short-cut Path}
\label{sec:shortcut}
The gating network determines how much each expert's anomaly score $\hat{y}_i$ contributes to the overall anomaly score $\hat{y}$.
We introduced a short-cut path to quickly shift the final anomaly score to 1, i.e. anomalous.
In other words, whenever the gating network knows that the input is anomalous, it can ignore the decision of the expert NNs.
In \Cref{fig:ablation}, we show the \textit{mean} gating weight for each expert NN in the example of MNIST.
Here, the gating network decided between 5~experts and the short-cut connection.
For normal data, i.e. data the gating network knows, the gating decision was approximately uniformly distributed reflecting the properties of the normal samples.
Looking at each test sample separately, we saw that the gating network assigned high importance to the expert NN of the respective class.
As expected, the short-cut path was not used as it should only be activated for anomalous samples.
Looking at the \textit{mean} gating weight for the anomalous test data, we see two cases:
1) in the semi-supervised case the short-cut path was often used, however, 2) for the unsupervised case mostly the decision of expert~3 was trusted.
In the unsupervised case, \ARGUE{} only uses generated anomalies, i.e. Gaussian noise, as counterexamples.
Thus, the anomaly decision is done by the expert NNs.
In the semi-supervised case, however, a few anomalies are already known, such that the gating network relies on the short-cut connection.
Based on our analysis, we saw that the short-cut connection matches our expectation: it is used when the input clearly is anomalous -- otherwise the anomaly decision is a mixture of the expert models.

\paragraph{Multi-Expert Architecture}
In \ARGUE{}, we concurrently analyse the activations of multiple expert NNs for anomalous patterns.
Could we have stuck to \AAA{}, where the activations of a single AE are analysed?
We saw similar performance of both AD methods in our semi-supervised experiments, while \ARGUE{} took the lead in the mean score.
In unsupervised environments, however, ARGUE performed \percincrease{.62}{.92}better, improving the mean AUC from $62\%$ to $92\%$ as shown in \Cref{tab:unsu_results}.
\AAA{}'s results exhibit considerably more variance, which results in a volatile performance and thus more manual work.
ARGUE's improvements due to the multi-expert architecture are three-fold:
1) the gating network automatically decides which expert path is important to the AD decision, 2) the short-cut path allows to quickly shift the output for anomalies, 3) each expert NN focuses on the features inherent to one notion of normal.
All these changes allow ARGUE to be applied in unsupervised and semi-supervised environments, whereas \AAA{} is limited to the latter.

\section{Discussion and Future Work}
ARGUE comprises multiple NNs, each contributing to the overall anomaly score.
We hope to motivate future research to port our architecture to other and potentially heterogeneous data types.
Federated learning may profit from the distributed knowledge among the expert networks.
We saw ARGUE's performance under three different clustering strategies.
Even without ideal clusters, we saw strong performance by using attributes already available in the input data.
Future research on clustering algorithms may find a way to bridge the gap between the presented self-supervised and unsupervised performance.
We hope to spark interest in AD research where ``normal'' is reconsidered under multiple contexts.

\section{Conclusion}
We introduced ARGUE: an anomaly detection method fusing the knowledge of multiple expert deep neural networks applicable to semi-supervised, self-supervised and unsupervised scenarios.
Each expert learns the distribution of parts of the training data, which are then analysed for anomalous patterns.
A gating mechanism weights the importance of each AD decision and fuses them to one overall anomaly score.
Our evaluation motivated that knowledge about the distribution of the normal data may be as valuable as known anomalies.
\ARGUE{} flexibly uses both types of prior knowledge, i.e. anomaly labels and data clusters, to boost the detection performance, yet also works in completely unsupervised settings.

\bibliographystyle{abbrv}
\bibliography{references}

\end{document}

%% file: figs/high_level.tikz
\begin{tikzpicture}[node distance=1.1*\NodeSepX cm]
	\node[smallannot] (norm_in) {\includegraphics[width=\textwidth]{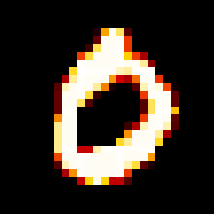}};

	\node[rect, right of=norm_in] (norm_exp1) {Expert for 0s};
	\node[rect, below of=norm_exp1, yshift=.75*\NodeSepY cm] (norm_exp2) {Expert for 1s};

	\node[smallannot, right of=norm_exp1] (norm_out1) {\includegraphics[width=\textwidth]{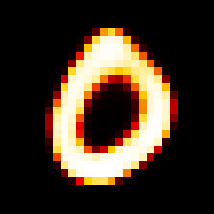}};
	\node[smallannot, right of=norm_exp2] (norm_out2) {\includegraphics[width=\textwidth]{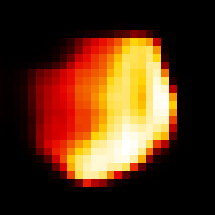}};

	\node[annot, below right of=norm_out1, text width=3*\NodeSepX cm, xshift=\NodeSepX cm, yshift=.7*\NodeSepY cm] (norm_annot) {Input familiar to the first expert, thus likely \emph{normal}.};

	\draw[arrow] (norm_in) -- (norm_exp1);
	\draw[arrow] (norm_in) |- (norm_exp2);
	\draw[arrow] (norm_exp1) -- (norm_out1);
	\draw[arrow] (norm_exp2) -- (norm_out2);

	\draw[line, dashed] (0,-1.8) -- (7.5,-1.8);

	\node[smallannot, below of=norm_in, yshift=-1.1*\NodeSepY cm] (anom_in) {\includegraphics[width=\textwidth]{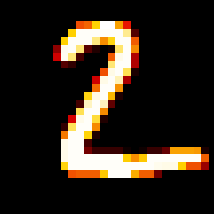}};

	\node[rect, right of=anom_in] (anom_exp1) {Expert for 0s};
	\node[rect, below of=anom_exp1, yshift=.75*\NodeSepY cm] (anom_exp2) {Expert for 1s};

	\node[smallannot, right of=anom_exp1] (anom_out1) {\includegraphics[width=\textwidth]{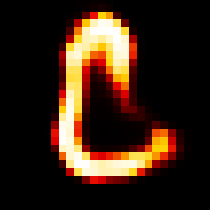}};
	\node[smallannot, right of=anom_exp2] (anom_out2) {\includegraphics[width=\textwidth]{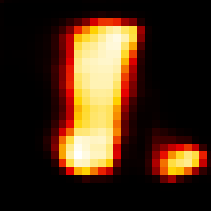}};

	\node[annot, below right of=anom_out1, text width=3*\NodeSepX cm, xshift=\NodeSepX cm, yshift=.7*\NodeSepY cm] (anom_annot) {Input not familiar to any expert, thus likely \emph{anomalous}.};

	\draw[arrow] (anom_in) -- (anom_exp1);
	\draw[arrow] (anom_in) |- (anom_exp2);
	\draw[arrow] (anom_exp1) -- (anom_out1);
	\draw[arrow] (anom_exp2) -- (anom_out2);

\end{tikzpicture}

%% file: figs/architecture_overview.tikz
\pgfdeclarelayer{bg} 
\pgfsetlayers{bg,main}

\begin{tikzpicture}[node distance=1.44 * \NodeSepX cm]
	\node[annot, text width=.5*\NodeSepX cm] (x) {$\vect{x}$};

	\node[rect, right of=x, xshift=-0.4*\NodeSepX cm] (enc) {Encoder};
	\draw[arrow] (x) |- (enc);

	\node[rect, right of=enc] (dec1) {Expert 1};
	\node[rect, below of=dec1, yshift=.5*\NodeSepY cm] (dec2) {Expert 2};
	
	\draw[arrow] (enc) -- (dec1);
	\draw[arrow] (enc) -- +(.75*\NodeSepX cm, 0)|- (dec2);

	\node[annot, gray, right of=dec1, text width=.5*\NodeSepX cm, xshift=-0.4*\NodeSepX cm] (xt1) {$\hat{\vect{x}}_1$};
	\node[annot, gray, right of=dec2, text width=.5*\NodeSepX cm, xshift=-0.4*\NodeSepX cm] (xt2) {$\hat{\vect{x}}_2$};

	\draw[arrow, dotted] (dec1) -- (xt1);
	\draw[arrow, dotted] (dec2) -- (xt2);

	\node[rect, below right of=dec1, yshift=\NodeSepY cm] (alarm1) {Alarm};
	\node[rect, below right of=dec2, yshift=\NodeSepY cm] (alarm2) {Alarm};

	\draw[arrow] (dec1.270) |- node[tinyannot, text width=\NodeSepX cm] {Hidden \\ Activation} (alarm1.178);
	\draw[arrow] (dec2) |- node[tinyannot, text width=\NodeSepX cm] {Hidden \\ Activation} (alarm2.178);
	\draw[arrow] (enc) |- (alarm1.182);
	\draw[arrow] (enc) |- (alarm2.182);

	\node[rect, below right of = enc, yshift=-3*\NodeSepY cm] (gating) {Gating};
	\draw[arrow] (enc.270) |- node[tinyannot, text width=\NodeSepX cm] {Hidden \\ Activation} (gating);

	\node[circle, tinyannot, right of=alarm1, draw=black, xshift=-.5*\NodeSepX cm] (mult_1) {};
	\node[cross out, minimum size=.2*\NodeSepX cm, draw=black] (mult_1_x)  at (mult_1) {};
	\node[circle, tinyannot, right of=alarm2, draw=black, fill=white, xshift=-.5*\NodeSepX cm] (mult_2) {};
	\node[cross out, minimum size=.2*\NodeSepX cm, draw=black] (mult_2_x)  at (mult_2) {};

	\draw[arrow] (alarm1) -- node[annot, above] {$\hat{y}_1$} (mult_1);
	\draw[arrow] (alarm2) -- node[annot, above] {$\hat{y}_2$} (mult_2);
	
	\begin{pgfonlayer}{bg}
		\draw[arrow] (gating) -| node[annot, near start, above] {$\hat{\vect{p}}$} (mult_1.280);
		\draw[arrow] (gating) -| (mult_2.270);
	\end{pgfonlayer}

	\node[circle, tinyannot, right of=gating, draw=black, xshift=1.3*\NodeSepX cm] (add) {};
	\node[cross out, minimum size=.2*\NodeSepX cm, draw=black, rotate=45] (add_x)  at (add) {};

	\draw[arrow] (mult_1)  -| (add.80);
	\draw[arrow] (mult_2) -| (add.100);
	\draw[arrow] (gating) -- (add);

	\node[annot, right of=add, xshift=-.8*\NodeSepX cm, text width=.5*\NodeSepX cm] (y) {$\hat{y}$};

	\draw[arrow] (add)  -- (y);
	
\end{tikzpicture}

%% file: figs/motivation_plot.tikz
\pgfplotstableread[col sep=comma]{csv/n_anomalies.csv}\data
\pgfplotsset{error_p/.style={error bars/.cd,	y dir=both,y explicit relative}}

\begin{tikzpicture}
	\begin{axis}[
        height=3.5cm,
		width=\linewidth,
		symbolic x coords={A,A-B,A-C,A-D,A-E,A-F,A-G,A-H},
		legend style={at={(0.45, -0.5)},anchor=north, legend columns=4, nodes={scale=0.7, transform shape}},
		grid=both,
		xtick=data,
		xlabel={Train Classes},
		ylabel={Mean AUC},
	]
		\addplot+[error_p] table[x=Exp.,y=ARGUE-AUC-mean,y error=ARGUE-AUC-std] {\data};
		\addplot+[error_p] table[x=Exp.,y=AE-AUC-mean,y error=AE-AUC-std] {\data};
		\addplot+[error_p] table[x=Exp.,y=MEx_CVAEC-AUC-mean,y error=MEx_CVAEC-AUC-std] {\data};
		\addplot+[error_p] table[x=Exp.,y=DAGMM-AUC-mean,y error=DAGMM-AUC-std] {\data};
		\addplot+[error_p] table[x=Exp.,y=GANomaly-AUC-mean,y error=GANomaly-AUC-std] {\data};
		\addplot+[error_p] table[x=Exp.,y=DeepSVDD-AUC-mean,y error=DeepSVDD-AUC-std] {\data};
		\addplot+[error_p] table[x=Exp.,y=REPEN-AUC-mean,y error=REPEN-AUC-std] {\data};
		\addplot+[error_p] table[x=Exp.,y=A3-AUC-mean,y error=A3-AUC-std] {\data};
		\legend{ARGUE, AE, MEx-CVAEC, DAGMM, GANomaly, DeepSVDD, REPEN, A3}
	\end{axis}
\end{tikzpicture}

%% file: figs/ablation_gate.tikz
\pgfplotstableset{col sep=comma}

\pgfplotstableread{csv/MNIST_gate.csv}\mnist

\begin{tikzpicture}
	\begin{axis}[
		height=3.5cm,
		width=\linewidth,
		legend style={at={(0.49, -0.35)},anchor=north, legend columns=-1},
		xtick=data,
		xticklabels from table={\mnist}{type},
		ybar, bar width=5pt,
		ymajorgrids,
		xmin=-0.5, xmax=3.5,
		ylabel={Mean Gate, $\bar{\hat{\vect{p}}}$},
	]
    \addplot[color=f1,fill] table[x expr=\coordindex, y={0}] {\mnist};
	\addplot[color=f1!80,fill] table[x expr=\coordindex, y={1}] {\mnist};

    \addplot[color=f1!60,fill] table[x expr=\coordindex, y={2}] {\mnist};
	\addplot[color=f1!40,fill] table[x expr=\coordindex, y={3}] {\mnist};

    \addplot[color=f1!20,fill] table[x expr=\coordindex, y={4}] {\mnist};
	\addplot[color=red,fill] table[x expr=\coordindex, y={5}] {\mnist};

	\legend{Ex. 0, Ex. 1, Ex. 2, Ex. 3, Ex. 4, Short-cut}

	\node [annot] at (axis cs:.5,.4) {Unsupervised};
	\node [annot] at (axis cs:2.5,.4) {Semi-supervised};

	\end{axis}
\end{tikzpicture}